\documentclass{article}

\usepackage{microtype}
\usepackage{graphicx}
\usepackage{subfigure}
\usepackage{booktabs} % for professional tables
\usepackage{amssymb}
\usepackage{amsthm}
\usepackage{amsmath}
\usepackage[utf8]{inputenc} % allow utf-8 input
\usepackage[T1]{fontenc}    % use 8-bit T1 fonts
\usepackage{booktabs}       % professional-quality tables
\usepackage{amsfonts}       % blackboard math symbols
\usepackage{nicefrac}       % compact symbols for 1/2, etc.
\usepackage{microtype}      % microtypography
\usepackage{xcolor}         % colors
\usepackage{bm}
\usepackage{enumitem}
\usepackage{mathbbol}
\newcommand{\norm}[1]{\left\lVert#1\right\rVert}
\DeclareMathOperator{\tr}{tr}
\include{figs}
\newtheorem{definition}{Definition}
\newtheorem{theorem}{Theorem}
\usepackage{hyperref}

\usepackage[accepted]{mlsys2025}

\mlsystitlerunning{ Private, Augmentation-Robust and Task-Agnostic  Data Valuation Approach for Data Marketplace}

\begin{document}

\twocolumn[
\mlsystitle{ Private, Augmentation-Robust and Task-Agnostic  Data Valuation Approach for Data Marketplace}

% unless you've provided the [accepted] option to the mlsys2024


\mlsyssetsymbol{equal}{*}

\begin{mlsysauthorlist}
\mlsysauthor{Tayyebeh Jahani-Nezhad}{ber}
\mlsysauthor{Parsa Moradi}{minn}
\mlsysauthor{Mohammad Ali Maddah-Ali}{minn}
\mlsysauthor{Giuseppe Caire}{ber}
% \mlsysauthor{Fiuea Rrrr}{to}
% \mlsysauthor{Tateu H.~Yasehe}{ed,to,goo}
% \mlsysauthor{Aaoeu Iasoh}{goo}
% \mlsysauthor{Buiui Eueu}{ed}
% \mlsysauthor{Aeuia Zzzz}{ed}
% \mlsysauthor{Bieea C.~Yyyy}{to,goo}
% \mlsysauthor{Teoau Xxxx}{ed}
% \mlsysauthor{Eee Pppp}{ed}
\end{mlsysauthorlist}

\mlsysaffiliation{ber}{Technische Universit{\"a}t Berlin, Berlin, Germany}
\mlsysaffiliation{minn}{University of Minnesota Twin Cities, Minneapolis, MN, USA}
% \mlsyskeywords{Machine Learning, MLSys}

\vskip 0.3in

\begin{abstract}
Evaluating datasets in data marketplaces, where the buyer aim to purchase valuable data, is a critical challenge. In this paper, we introduce an innovative task-agnostic data valuation method called \textit{$\mathsf{PriArTa}$} which is an approach for computing the distance between the distribution of the buyer's existing dataset and the seller's dataset, allowing the buyer to determine how effectively the new data can enhance its dataset. \textit{$\mathsf{PriArTa}$} is communication-efficient, enabling the buyer to evaluate datasets without needing access to the entire dataset from each seller. Instead, the buyer requests that sellers perform specific preprocessing on their data and then send back the results. Using this information and a scoring metric, the buyer can evaluate the dataset. The preprocessing is designed to allow the buyer to compute the score while preserving the privacy of each seller's dataset, mitigating the risk of information leakage before the purchase. A key feature of $\mathsf{PriArTa}$ is its robustness to common data transformations, ensuring consistent value assessment and reducing the risk of purchasing redundant data. The effectiveness of $\mathsf{PriArTa}$ is demonstrated through experiments on real-world image datasets, showing its ability to perform privacy-preserving, augmentation-robust data valuation in data marketplaces.
\end{abstract}
]



\printAffiliationsAndNotice{}  % leave blank if no need to mention equal contribution
% \printAffiliationsAndNotice{\mlsysEqualContribution} % otherwise use the standard text.

\section{Introduction}
The availability of large and relevant datasets has been essential for achieving high-performance machine learning models 
\cite{sun2017revisiting}. However, in critical fields like medical research, access to data is often severely restricted.  As a result, we have to purchase the data from marketplaces~\cite{agarwal2019marketplace}.
% Consider a scenario in which an entity, referred to as a data buyer, aims to perform a learning task. However, the quantity of the available dataset may be insufficient to meet the training requirements. In such instances, the data buyer may consider purchasing additional datasets from entities known as data sellers. A crucial factor to consider in this process is the value of the dataset the buyer intends to acquire and its impact on the training process. 
In a data marketplace, a buyer seeks to purchase data from various sellers. A key challenge for the buyer is evaluating each dataset before making a purchase, as varying data quality can significantly impact model performance in real-world applications. This raises a fundamental question: how can we quantify the value of data? There has been a wide range of work on data valuation from various perspectives and for different use cases. The methods can be categorized into two main types: \emph{task-oriented} and \emph{task-agnostic}.

Task-oriented methods are specifically designed to calculate the value of data for a given learning task. In \cite{ghorbani2019data}, $\mathsf{Data\ Shapley}$, inspired by the Shapley value in game theory \cite{shapley1953value}, is proposed to quantify the importance of each individual data point in a dataset by evaluating how much it contributes to the model’s performance. This is achieved by considering the difference in validation performance when including and excluding the data point across all subsets of the training dataset.  Other related works, such as $\mathsf{KNN-Shapley}$~\cite{jia2019efficient}, $\mathsf{Distributional\ Shapley}$ \cite{ghorbani2020distributional},  $\mathsf{Beta\ Shapley}$ \cite{kwon2021beta}, and $\mathsf{Bahnzhaf\ value}$ \cite{wang2023data} have been proposed to improve the Data Shapley value method in terms of computational cost or to generalize it further. The mentioned task-oriented data valuation methods are computationally expensive, as the performance must be calculated for each \textit{individual} data point. 
This makes them impractical for real-world applications, especially for large and complex models.

The second category of evaluations is task-agnostic methods, which do not depend on any model performance. Still, some task-agnostic methods such as $\mathsf{LAVA}$ \cite{just2023lava} and $\mathsf{DAVINZ}$ \cite{wu2022davinz} calculates the value of each \textit{individual} data point in the dataset. For example, $\mathsf{LAVA}$ calculates the contribution of each point by measuring the gradient of a defined distance between the training set and a validation set with respect to the probability mass of that point. $\mathsf{DAVINZ}$ derives a domain-aware generalization bound using the neural tangent kernel (NTK) \cite{jacot2018neural} to characterize and estimate the validation performance of a deep neural network without model training and uses this as the scoring function in the conventional data valuation technique in \cite{cook1977detection}. In \cite{nohyun2022data}, 
the Complexity-gap Score (CG-score) is introduced,  which is a training-free method for quantifying the impact of individual data points. CG-score leverages the properties of the Gram matrix in over-parameterized neural networks, which can be  efficiently computed from the dataset. In \cite{tay2022incentivizing}, a task-agnostic data valuation is used in collaborative machine learning, a type of ML that encourages self-interested parties to contribute data to a shared pool of training dataset. In \cite{tay2022incentivizing}, the data is valued based on its quantity and quality, measured by its closeness to the ground truth distribution using  Maximum Mean Discrepancy (MMD) distance \cite{gretton2012kernel}.
All of these approaches assume a centralized setting in which all data is fully accessible to measure the value of each data point, limiting their application for large datasets in data marketplaces.  In data marketplaces, often sellers do not want to send their raw data to the buyer due to privacy concerns and the risk of information leakage. The sellers prefer to keep their datasets private until a purchase is made. Furthermore, in some cases, sending all data from every seller to a centralized entity, such as the buyer, is not feasible due to communication load.

% One of the most important applications of data valuation is in the data marketplace. Consider a scenario in which an entity, referred to as a data buyer, aims to perform a learning task. However, the quantity of the available dataset may be insufficient to meet the training requirements. In such instances, the data buyer may consider purchasing additional datasets from entities known as data sellers. A crucial factor to consider in this process is the value of the dataset the buyer intends to acquire and its impact on the training process. 
In \cite{amiri2023fundamentals}, a task-agnostic data valuation method specifically designed for data marketplaces is proposed, where two metrics, \textit{diversity} and \textit{relevance}, are defined for sellers' datasets. These metrics are calculated by comparing the statistical properties of the seller's dataset with those of the buyer's dataset, measuring the difference and similarity based on the principal component space of the buyer's data. An extended version of that work is presented in \cite{luprivate}, which introduces alternative definitions for the diversity and relevance metrics and evaluates them on various image datasets. Although \cite{amiri2023fundamentals} and \cite{luprivate} do not provide a unique score for the value of each seller's dataset or a decision-making process for seller selection, a key drawback is that the defined metrics are not robust against natural or intentional augmentations that the sellers' datasets may undergo.

 % Consider a scenario in which an entity, which we call a data buyer, aims to perform a learning task. However, the quantity of the dataset available to meet the training requirements may be insufficient. In such instances, the option of purchasing additional datasets from entities commonly referred to as data sellers may be considered. The main important point that becomes a crucial factor to consider is the value of the data the buyer intends to acquire and its impact on the training process.

 % For example, suppose the objective is to train a neural network for diagnosing specific diseases, and there is a need for healthcare datasets sourced from various hospitals. In this scenario, the value of the data the buyer intends to acquire and its impact on the training process become crucial factors to consider. .......

In this paper, we introduce an innovative task-agnostic framework to evaluating sellers' datasets in a data marketplace, where a buyer aims to enhance its current dataset by purchasing datasets that offer the most significant value and coverage of the target population. Our framework, \textit{$\mathsf{PriArTa}$}, measures the distance between the distribution of the buyer's existing dataset and the seller's dataset. The value of a new dataset to a buyer depends on their specific preferences—whether they aim to enrich existing areas of their dataset or to cover underrepresented domains. Accordingly, the buyer can construct a utility function based on various dataset parameters. 
Our approach, \textit{$\mathsf{PriArTa}$}, has the following properties:
\begin{itemize}
    \item  $\mathsf{PriArTa}$ is designed to evaluate entire datasets, rather than individual data points, making it computationally efficient even for large-scale datasets.
    \item  One of the key strengths of $\mathsf{PriArTa}$ is its robustness to common data transformations. By ensuring that the value assigned to a dataset remains consistent even when the data has undergone transformations such as rotation, resizing, cropping, or color adjustments, $\mathsf{PriArTa}$ prevents the purchase of seemingly valuable datasets that cover different domains, and focuses on acquiring genuinely novel and beneficial data.
       \item  
       % Unlike the most of existing methods that require access to raw data, 
       $\mathsf{PriArTa}$ allows buyers to evaluate the value of sellers' datasets without needing direct access to the raw data. This approach ensures the privacy of sellers by allowing them to share information about their datasets after preprocessing and applying noise masking.
\end{itemize}
% \moh{To improve computational efficiency of estimating the empirical distance between the distributions, $\mathsf{PriArTa}$ benefits from variational Bayesian methods \cite{kingma2013auto,rezende2014stochastic} to transform the distribution of the datasets to Gaussian distributions. Such distributions are parameterised by their mean and covariance matrices, which can be empirically computed, and used to compute the distance. Leveraging concepts from contrastive learning \cite{chen2020simple}, we make  $\mathsf{PriArTa}$ data evaluation robust against augmentation. By adding noise XXXX, we make $\mathsf{PriArTa}$  differential private~\cite{dwork2014algorithmic,dwork2006calibrating}.}

% The metric proposed in $\mathsf{PriArTa}$ enables the buyer to assign a value score to each seller's dataset in a privacy-preserving manner. This metric is designed so that the value of a dataset and its transformed versions does not differ significantly, helping to avoid incorrect decision-making in seller selection. 
By leveraging concepts from contrastive learning \cite{chen2020simple}, variational Bayesian methods \cite{kingma2013auto}, statistical distances \cite{kantorovich1942translocation}, and differential privacy \cite{dwork2014algorithmic,dwork2006calibrating}, our method provides a privacy-preserving, augmentation-robust, and communication-efficient solution for data valuation in data marketplaces. Experiments on real-world image datasets demonstrate the effectiveness of $\mathsf{PriArTa}$, even when sellers possess augmented versions of other sellers' or the buyer's datasets.

\section{Problem Formulation}
Consider a distributed data marketplace framework comprising one buyer holding a dataset $D_{B}$ consisting of $m_0$ data points, 
each sampled i.i.d. from an unknown distribution $P_{B}$, represented as $D_{B} \sim P_B$. Additionally, there are $N$ sellers, wherein each seller $i \in {1, \dots, N}$ holds a dataset $D_{S_i}$ consisting of $m_i$ data points,  each of size $L$. Analogous to the buyer's dataset, every data point within these datasets is sampled i.i.d. from an unknown distribution, denoted as $D_{S_i} \sim P_{S_i}$. 
% We assume the existence of an underlying data distribution $P^{*}$, from which all distributions originate, although they do not necessarily have identical statistics. 
Suppose the buyer, irrespective of any specific machine learning task, defines a score function $u(D_B; D_{S_i})$ to evaluate each seller's dataset relative to its own dataset $D_B$. 
By selecting the seller’s dataset with the greatest score or dissimilarity, the buyer can enhance coverage of the target domain; alternatively, choosing the dataset with lower dissimilarity allows the buyer to enrich their existing data  in the areas the buyer already has data. In evaluating $u(D_B; D_{S_i})$, we aim to satisfy specific properties, as explained below: 

% \textcolor{red}{The buyer aims to find a dataset that maximizes dissimilarity to its own, thereby enhancing coverage of the target domain and improving task performance afterwards.}

{\bf Communication Efficiency:} In data marketplaces, it is not desirable for the sellers to send the entire datasets to the buyer before making a purchase. On the other hand, for the buyer, receiving and processing all datasets is not feasible. Instead, the buyer should calculate the score function based on specific information requested from each seller. 
% This approach allows the buyer to assess the value of the seller's dataset without communicating the entire datasets. 
Each seller performs preprocessing on a subset of its dataset  using a  mapping function $f_{\text{map}}(\cdot)$, which maps a subset of size $\tilde{m}_i\le m_i$ of  dataset $D_{S_i}$ to a matrix $\mathbf{X}_{S_i}\in \mathbb{R}^{b\times q}$ as the representation of its dataset, where $b$ and $q$ are integer numbers and $bq\lll \tilde{m}_i L$. Note that this subset of data samples is chosen at random from the entire dataset, ensuring unbiased representation.

{\bf Privacy:} To ensure that the individual datasets of the sellers remain private, each seller should mask the representation $\mathbf{X}_{S_i}$ using some noises and generate $\tilde{\mathbf{X}}_{S_i}$ so that it cannot be inferred by the buyer. Then each seller $i$ shares the masked representation $\tilde{\mathbf{X}}_{S_i}$ to the buyer so that it can calculate the score function.

Consider the buyer's score function applied to the dataset of seller-$i$ is denoted by $u(D_B; D_{S_i}) \triangleq d(\mathbf{X}_B, \tilde{\mathbf{X}}_{S_i})$, where $d(\cdot)$ represents a distance metric, and $\mathbf{X}_B\triangleq f_{\text{map}}(D_B)$.
Consequently, the buyer chooses to purchase the dataset from the seller that best aligns with their specific needs, either by covering underrepresented domains or by enriching data within existing areas.

% \textcolor{red}{Consequently, the buyer chooses to purchase the dataset from the seller with the highest score value.} More precisely, ${S^{*}} \triangleq \arg\max_{i \in \{1,\dots,N\}} d(\mathbf{X}_B, \mathbf{X}_{S_i})$, where ${S^{*}} $ denotes the selected seller for priority purchase.

{\bf Transformation Resistance:} To ensure the buyer acquires only unique datasets with minimum redundancy, it's essential to avoid purchasing datasets that are  transformations of other sellers' datasets or even the buyer's. This includes datasets that have been altered in ways such as rotation, cropping, resizing, color adjustments, or any similar augmentations.
Therefore, the combination of the mapping function $f_{\text{map}}(\cdot)$ and the distance metric $d(\cdot)$ should be resistant to these kinds of transformations. Consider $\tilde{x}_j\triangleq t(x_j)$, where $\tilde{x}_j$ represents the augmented version of data sample $x_j\in D_{S_i}$, and $t(\cdot)$ denotes a function randomly applying various transformations from a predetermined set. In this scenario, the distance metric should satisfy the condition:
\begin{align*}
    \left| d\big(f_{\text{map}}(D_{B}), f_{\text{map}}(D_{S_{i}})\big) - d\big(f_{\text{map}}(D_{B}), f_{\text{map}}(\tilde{D}_{S_{i}})\big) \right| \leq \varepsilon,
\end{align*}
where $\varepsilon$ is a small positive value representing the acceptable difference in distances, and $\tilde{D}_{S_{i}}\triangleq\{t(x_j), \forall x_j\in D_{S_i}\}$. This condition ensures that the discrepancy between the original distance and the distance to the transformed dataset is bounded by $\varepsilon$, indicating that the two distances  do not significantly differ.
\subsection{Our Contribution} In this paper,  we propose $\mathsf{PriArTa}$ framework, a private, task-agnostic, and augmentation-robust data valuation method that evaluates entire datasets, rather than individual data points.
The core idea behind $\mathsf{PriArTa}$ is driven by the need to measure the distance between the distributions of the buyer's and the sellers' datasets. Typically,  popular distance metrics for comparing distributions often require access to the actual distribution or an empirical estimate of it  \cite{gibbs2002choosing}, which can increase the complexity of these metrics, particularly in high-dimensional settings \cite{peyre2019computational}.
% However, estimating these distributions accurately can be computationally expensive due to high sample complexity \cite{}.

To address this, we propose mapping the distribution of each seller's dataset to a parametric distribution, which can be characterized by a few key parameters. One of the most widely used parametric distributions is the Gaussian distribution. If all the sellers' distributions are approximated by Gaussian distributions, we can then employ a proper distance metric such as the Wasserstein distance \citep{kantorovich1942translocation}, which offers a closed-form solution for computing the distance between two Gaussian distributions based on their first and second moments, i.e., the mean and covariance.

To achieve this mapping, we suggest using a Variational Autoencoder (VAE) \citep{kingma2013auto}, which allows the latent variable to approximate a Gaussian distribution given an input dataset. Additionally, to ensure that the mapping is robust against data augmentations, we propose utilizing a SimCLR model \cite{chen2020simple} which generates representations of the dataset that are invariant to transformations, thus making the subsequent distance measurement more reliable. Furthermore, to preserve the privacy of the sellers' datasets, $\mathsf{PriArTa}$ employs local differential privacy \cite{dwork2014algorithmic,dwork2006calibrating} with the Gaussian mechanism.

\section{Building Blocks of $\mathsf{PriArTa}$}
In this section, we provide a high-level overview of the $\mathsf{PriArTa}$ framework, describing its main building blocks, followed by an explanation of a specific implementation for each block. The detailed version of the proposed private, task-agnostic, and augmentation-robust data valuation framework will be described in the next section.  $\mathsf{PriArTa}$ is composed of modular components, each addressing essential challenges in data valuation for marketplaces, as previously discussed: (Mod.1) unsupervised representation learning,  (Mod.2) distribution mapping, (Mod.3) valuation metrics, and  (Mod.4) privacy-preserving mechanisms. This modular structure provides flexibility in method selection for each component, allowing the data valuation process to be tailored to specific requirements.

\subsection{Mod.1: Unsupervised Representation Learning} This module focuses on generating data representations that capture underlying patterns in the dataset while maintaining robustness against common transformations, such as cropping, resizing, flipping, color jittering, and Gaussian blur, which may be applied in combination.
Here, robustness refers to the learned representation’s invariance to augmentations applied to the input dataset.
Within this framework, this module aids the data valuation framework in performing reliable valuations on datasets, regardless of potential visual augmentations, ensuring consistency in data valuation outcomes.

One option for this module is contrastive learning methods, which lead the field in self-supervised learning. The objective is to learn a meaningful data representation without relying on explicit supervision or labels by contrasting positive and negative pairs of instances, where similar data points are positioned close together in the representation space, and dissimilar ones are positioned farther apart \citep{bachman2019learning}.
Among the various contrastive learning methods suitable for implementing $\mathsf{PriArTa}$, we select SimCLR \citep{chen2020simple} as a straightforward example.
In SimCLR, multiple augmented versions of the same instance are treated as positive pairs, while different samples serve as negative pairs. The model’s objective function is to differentiate between these positive and negative pairs to capture meaningful and semantic information within the data. SimCLR consists of several components, including:

(1) A stochastic data augmentation module that randomly transforms input data $\mathbf{x}$, producing two augmented versions $\tilde{\mathbf{x}}_i$ and $\tilde{\mathbf{x}}_j$ of the same instance, which are considered as positive pairs. Common data augmentation techniques used include cropping, flipping, rotation, random crop, and color transformations.

(2) An encoder network $f_{\text{SimCLR}}(\cdot)$, typically a deep neural network architecture like ResNet \cite{he2016deep}, takes the augmented instances and extracts the representation vectors, making the discrimination between similar and dissimilar instances higher in cooperation with the other components. Let us denote the output of the encoder by $\mathbf{h}_i=f_{\text{SimCLR}}(\tilde{\mathbf{x}}_i)$.

(3) A projection head $g_{\text{head}}(\cdot)$ is employed to further refine the learned representation, which is a shallow neural network like a Multilayer Perceptron (MLP) \cite{rumelhart1986learning}. It maps the representations to a lower-dimensional space where contrastive loss is applied. The output of this projection head is denoted by $\bm{\nu}_i=g_{\text{head}}(\mathbf{h}_i)$, representing the learned representation with more discriminative power.

(4) A contrastive loss is defined to be applied to the encoded and projected instances, aiming to bring similar instances closer together and push dissimilar instances apart. To achieve this goal, a distance metric such as Euclidean distance or cosine similarity is required.

In order to train the model, a minibatch consisting of $N$ data samples is randomly chosen, and contrastive learning is performed on pairs of $2N$ augmented versions of data samples derived from the minibatch. Negative samples are not explicitly chosen. However, for each augmented pair, the other $2(N-1)$ augmented samples within the minibatch are considered as negative data samples. Considering the cosine similarity as a distance metric, the loss function, called the normalized temperature-scaled cross-entropy loss (NT-Xnet), for a positive pair $(i,j)$ in SimCLR is defined as follows.
\begin{align*}
        \ell_{i,j}=-\log{\frac{\exp{(s_{i,j}/\tau)}}{\sum_{k=1}^{2N}\mathbb{1}_{k\ne i}\exp{(s_{i,k}/\tau)}}},
\end{align*}
where $s_{i,j}$ is a cosine similarity function defined as $s_{i,j}={\bm{\nu}_i^T\bm{\nu}_j}/({\norm{\bm{\nu}_i}\norm{\bm{\nu}_j}})$, $\mathbb{1}_{k\ne i}$ is an indicator function and $\tau$ denotes a temperature parameter. The last step is to calculate and sum all the losses as 
\begin{align*}
    \mathcal{L}=\frac{1}{2N}\sum_{k=1}^{N}[\ell_{2k-1,2k}+\ell_{2k,2k-1}],
\end{align*}
and update the encoder network and the projection head such that $\mathcal{L}$ is minimized. Learned representations from SimCLR can be transferred to downstream tasks by extracting the representations from the output of the encoder network of the SimCLR model. A decision should be made on whether to fine-tune the SimCLR model or use its fixed features as input to a new model.

Note that for this module, methods other than SimCLR, including other contrastive learning or even non-contrastive learning methods that meet the specified requirements and align with $\mathsf{PriArTa}$ framework, can also be used.
\subsection{Mod.2: Distribution Mapping}
The main idea of $\mathsf{PriArTa}$ is to evaluate datasets by measuring the distance between the buyer’s and seller’s dataset distributions. Popular metrics for comparing distributions often require access to full empirical estimates of these distributions, which can be computationally intensive, particularly in high-dimensional scenarios. This module addresses this challenge by transforming datasets into structured representations that simplify the process of comparing distributions.

The objective of this module is to map the distribution of each dataset into a manageable and parametric format. In particular, a parametric distribution can be characterized by a limited set of parameters that enable efficient and meaningful comparisons between datasets. Among the options, Gaussian distributions are commonly employed due to their simplicity and ease of analysis. This choice also supports computationally efficient distance metrics and provides a structured approach to represent each dataset's statistical characteristics, ensuring that the framework can scale with large datasets. For this module, we select the Variational Autoencoder (VAE) \citep{kingma2013auto} due to its ability to learn compact latent representations of complex data and impose a Gaussian structure on the latent space.

A VAE is a type of generative model that learns to represent and generate data in a probabilistic manner.
It is composed of an encoder, which maps high-dimensional input data to a lower-dimensional latent space, and a decoder, which maps points in the latent space back to the data space.  
The objective of a VAE includes two terms: a reconstruction loss, which encourages the model to  reconstruct input data, and a regularization term, which encourages the latent variables to approximate the prior distribution, typically a Gaussian distribution. This regularization term, often implemented using Kullback-Leibler (KL) divergence \cite{kullback1951information}, helps prevent overfitting and encourages the latent space to have certain properties, such as continuity and smoothness.
VAEs aim to  maximize a lower bound on the log-marginal likelihood, which consists of the  reconstruction loss and the KL divergence between the encoder and the prior distribution. 
% Minimizing the reconstruction loss in isolation is equivalent to training a deterministic auto-encoder. For this reason, the rate is often interpreted as a regularizer that induces a smoother representation.
% using maximum likelihood estimation with a variational inference approach to approximate the posterior distribution of latent variables. 
More precisely, components of a VAE can be described as follows: \\
\textbf{Encoder}: The encoder is a neural network with datapoints $\mathbf{x}$ as the input data and the latent variable $\mathbf{z}$ as the output.  The encoder parameterizes the approximate variational posterior distribution $q_{\theta}(\mathbf{z}|\mathbf{x})$ of the latent variable $\mathbf{z}$ given the input data $\mathbf{x}$ and maps the input data to the parameters of a Gaussian distribution in the latent space. 
\begin{align*}
    q_{\theta}(\mathbf{z}|\mathbf{x}) = \mathcal{N}(\bm{\mu}(\mathbf{x}), \bm{\sigma}^2(\mathbf{x})),
\end{align*}
where $\theta$ denotes the parameters of the encoder model, $\bm{\mu}(\mathbf{x})$ and $\bm{\sigma}^2(\mathbf{x})$ are the mean and variance vectors computed by the encoder network respectively.
After approximating the variational posterior $q_{\theta}(\mathbf{z}|\mathbf{x})$, the latent variable $\mathbf{z}$ is obtained by sampling from this distribution.\\
\textbf{Decoder:}  The decoder is another neural network which maps the sampled latent variable $\mathbf{z}$ back to the data space to generate a reconstruction $\mathbf{x}'$. The decoder outputs the parameters to the likelihood function $p_{\phi}(\mathbf{x}|\mathbf{z})$, where $\phi$ denotes the parameter of the decoder model.\\
\textbf{Objective Function:}
The VAE is trained by maximizing the evidence lower bound (ELBO), which is the lower bound of the log-likelihood of the data as follows.
\begin{align*}
    \text{ELBO}(\mathbf{x}) = \mathbb{E}_{\mathbf{z}\sim q_{\theta}(\mathbf{z}|\mathbf{x})}[\log p_{\phi}(\mathbf{x}|\mathbf{z})] - \text{KL}(q_{\theta}(\mathbf{z}|\mathbf{x}) || p(\mathbf{z})),
\end{align*}
 where the first term of the objective is a reconstruction loss, and $\text{KL}(q_{\theta}(\mathbf{z}|\mathbf{x}) || p(\mathbf{z}))$ is the KL divergence between the approximate posterior and the prior distribution of the latent variable $\mathbf{z}$. It has been shown that the KL loss term,  induces a smoother representation of the data \citep{chen2016variational}. 
By maximizing the ELBO, the VAE learns to  extract the meaningful representation of the input data in the latent space, allowing it to generate new data samples that resemble the training data.

\subsection{Mod.3: Valuation Metric}
To quantify the value of each dataset, the framework requires a distance-based metric that compares the distributions of the buyer’s and seller’s data representations. Finding an effective distance metric is critical. 
 
In $\mathsf{PriArTa}$, we select optimal transport (OT) as a metric for computing the distance between two probability distributions which is both symmetric and satisfies the triangle inequality. The goal of optimal transport is to find the most efficient way to minimize the total cost of moving probability mass from a source distribution to a target distribution, subject to certain constraints. Let measures $\alpha$ and $\beta$ be probability distributions on spaces $\mathcal{X}$ and $\mathcal{Y}$ respectively, which can be continuous or discrete. Let $c:\mathcal{X}\times\mathcal{Y}\to [0,+\infty]$ be a symmetric cost function where $c(x,y)$ measures the cost of transporting one unit of mass from an element $x$ in $\mathcal{X}$ to an element $y$ in $\mathcal{Y}$ (with property $c(x,x)=0$). The optimal transport problem seeks to minimize the total cost of transporting mass from $\alpha$ to $\beta$ and it is defined as follows:
\begin{align*}
    \text{OT}(\alpha,\beta)=\min_{\pi\in\Pi(\alpha,\beta)}\int_{\mathcal{X}\times\mathcal{Y}}c(x,y)d\pi(x,y),
\end{align*}
where $\Pi(\alpha,\beta)$ is the set of couplings consisting of joint probability distributions over the space $\mathcal{X}\times\mathcal{Y}$ with $\alpha$ and $\beta$ as marginals. More precisely, 
\begin{align*}
    \Pi(\alpha,\beta)=\big\{\pi\in\mathcal{P}(\mathcal{X}\times\mathcal{Y})&\bigg| \int_{\mathcal{X}}\pi(x,y)dx=\beta, \\\nonumber
    &\int_{\mathcal{Y}}\pi(x,y)dy=\alpha\big\}.
\end{align*}
The $p$-Wasserstein distance is a metric derived from optimal transport, with the Euclidean distance serving as the cost function. It is defined as follows:
\begin{align*}
    W_p(\alpha,\beta)=\big( \min_{\pi\in\Pi(\alpha,\beta)}\int_{\mathcal{X}\times\mathcal{Y}}\norm{x-y}^p d\pi(x,y)\big)^{\frac{1}{p}},
\end{align*}
where $p\ge 1$. The $p$-Wasserstein distance provides a valuable tool for comparing and analyzing probability distributions in various fields, including statistics, machine learning, and image processing. A special case occurs for  multivariate normal distributions. If $\alpha=\mathcal{N}(\mu_1,\Sigma_1)$ and $\beta=\mathcal{N}(\mu_2,\Sigma_2)$, then the distance has a closed form as follows: 
\begin{align}
\label{Wasserstein}
    W_2(\alpha,\beta)=\big(&\norm{\mu_1-\mu_2}^2 + \tr(\Sigma_1)+\tr(\Sigma_2)\\\nonumber
    &-2\tr\big( (\Sigma_1^{1/2}\Sigma_2\Sigma_1^{1/2})^{1/2}\big)\big)^{1/2}.
\end{align} 
In $\mathsf{PriArTa}$, the $p$-Wasserstein distance is used for this module due to its simple analytical expression between two Gaussian distributions. Other distance metrics between distributions could also be considered, but they may encounter some challenges. Some commonly used metrics, such as KL divergence, do not satisfy symmetry and the triangle inequality and may even yield infinite values if the distributions do not share the same support. Maximum mean discrepancy distance is another metric, but its performance depends heavily on the choice of kernel, and it only considers the difference between the mean embeddings of the two distributions \cite{tay2022incentivizing}. 
% However, considering the required conditions, any suitable metric can be selected. 

\subsection{Mod.4: Privacy-Preserving Mechanism}
In a data marketplace, different types of privacy considerations may arise. One of the most important is the privacy of the sellers' datasets, as sellers might not wish to share their data before any purchases are made. This module provides sellers' privacy protection to minimize the risk of data leakage. 

Differential Privacy (DP) is a widely adopted privacy-preserving method in fields like data analysis, machine learning, and data sharing.
Local Differential Privacy (LDP), a variation of DP, is designed to ensure that individual data entries remain private even after they are collected or centralized. In the context of data privacy, LDP offers a robust framework for protecting users' information while still allowing meaningful statistical analysis on the collected data.
\begin{definition}
    A mechanism $\mathcal{M}:\mathcal{X}\to\mathcal{R}$ with data domain $\mathcal{X}$ and range $\mathcal{R}$ satisfies $(\epsilon,\delta)$-Local Differential Privacy \cite{dwork2006calibrating,kasiviswanathan2011can}, if for every pair of adjacent  inputs  $x_1, x_2 \in \mathcal{X}$  and for any possible output $O\subseteq \mathcal{R}$, the following inequality holds
    \begin{align*}
        Pr[\mathcal{M}(x_1)\in O]\le e^{\epsilon} Pr[\mathcal{M}(x_2)\in O]+\delta,
    \end{align*}
    where $\epsilon\ge 0$, $0\le \delta\le 1$,  
    and the probabilities are taken over the randomness of the mechanism. In addition, 
    two inputs are said to be adjacent if they differ in the data of exactly one individual.
\end{definition}
LDP ensures that the input to  $\mathcal{M}$ 
cannot be inferred from its output with high confidence, as determined by $\epsilon$. 
In this paper, we focus on the output perturbation LDP mechanism \cite{dwork2014algorithmic} which involves adding a random noise vector $Z$ to the output of a function 
$f:\mathcal{X}\to \mathbb{R}^d$. To ensure that the mechanism  $\mathcal{M}(x)=f(x)+\mathbf{n}$ meets the specific privacy guarantee, the noise level must be carefully calibrated based on the sensitivity of the function $f$ to input variations and the chosen noise distribution. The \textit{Gaussian mechanism} is employed to achieve this, where the perturbation $\mathbf{n}$ is modeled as an isotropic Gaussian noise vector with zero mean, i.e., $\mathbf{n}\sim\mathcal{N}(0,\sigma^2I)$, and the sensitivity of function $f$ is characterized by $\norm{f(x_1)-f(x_2)}_2\le \Delta$.
\begin{theorem}[Gaussian Mechanism]
 For $\epsilon,\delta\in(0,1)$,  the Gaussian mechanism with parameter $\sigma\ge c\Delta/\epsilon$ is $(\epsilon,\delta)$-differentially private, provided that $c^2>2\ln{(1.25/\delta)}$ \cite{dwork2014algorithmic,dwork2006calibrating}.
\end{theorem}

\section{Detailed Description of $\mathsf{PriArTa}$}\label{scheme}
In this section, we introduce $\mathsf{PriArTa}$ method for calculating the valuation of each dataset $D_{S_i}\sim P_{S_i}$, for $i=1, \dots, N$, where $N$ represents the number of sellers, with respect to the buyer's current dataset $D_B\sim P_B$. The proposed valuation method is task-agnostic, meaning that the value of each dataset remains unaffected regardless of the specific task the buyer intends to perform.
The main idea behind this method is to compute a distance metric between the buyer's dataset $D_B$ and each seller's dataset $D_{S_i}$. To achieve this goal, and to ensure to have a robust distance metric resilient to data augmentation,  the buyer and sellers engage in the following steps:

\begin{itemize}[leftmargin=*]
\item  As shown in Fig.~\ref{simclr_buyer}, the buyer initiates a contrastive learning algorithm on its dataset $D_B$, with the option to employ various self-supervised learning methodologies.
% Nevertheless, in our approach, we adopt SimCLR to train a deep neural network capable of capturing resilient representations of the input dataset. 
Specifically, in our framework, SimCLR is employed to extract representations that are invariant to data augmentations and to capture the underlying structure of the data through contrastive learning. In addition, SimCLR has a simple structure and strong performance, making it a suitable choice for our data marketplace framework.

% It is worth nothing that in $\mathsf{PriArTa}$ we assume that all sellers have datasets homogeneous with the buyer's dataset domain.
\begin{figure*}
  \centering
  % \fbox{\rule[-.5cm]{0cm}{4cm} \rule[-.5cm]{4cm}{0cm}}
  \includegraphics[width=1.3\columnwidth]{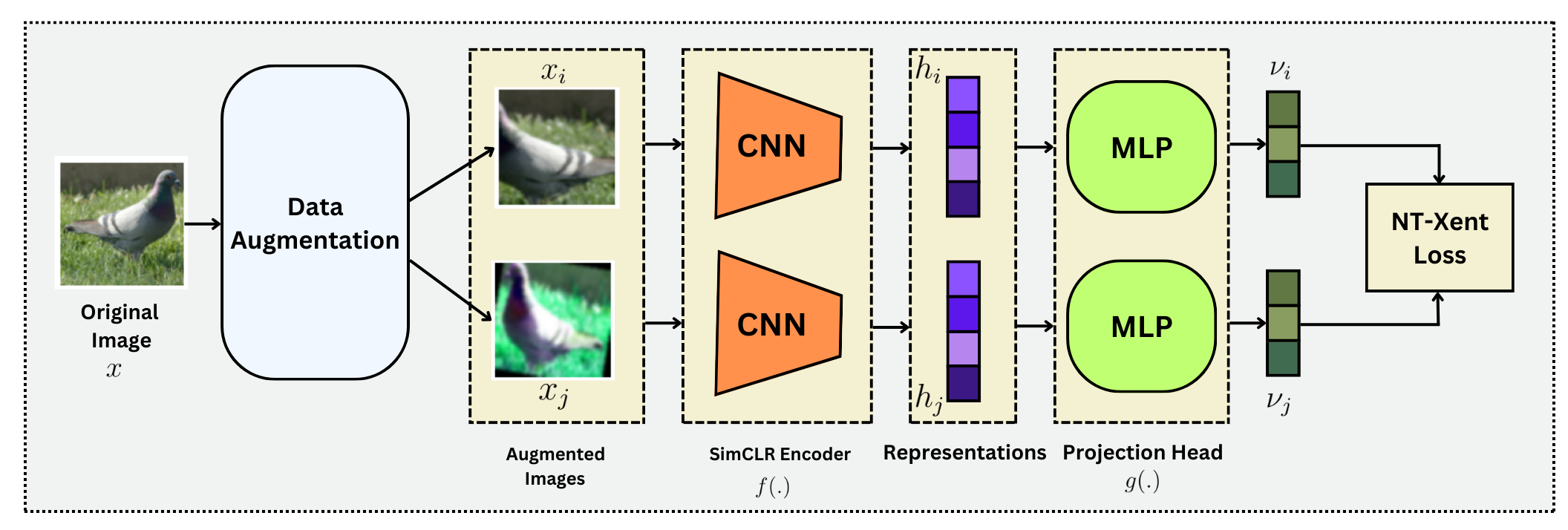}
  \caption{SimCLR training framework at the buyer side}
  \label{simclr_buyer}
\end{figure*}
\item 
The output of the backbone of the SimCLR model, denoted as $\mathbf{h}_B$, serves as the input to the encoder component of a variational auto-encoder which aims to minimize the deviation of the learned latent distribution from the standard normal prior distribution. 
The decoder also attempts to reconstruct these representations from sampled latent variables, $\mathbf{z}_B$. Subsequently, the buyer trains the VAE using its own datasets, as illustrated in Fig. \ref{vae_buyer}. Here, The VAE provides a principled way to regularize the latent space by enforcing a Gaussian prior, ensuring that the latent space is well-structured.
\begin{figure*}
  \centering
  % \fbox{\rule[-.5cm]{0cm}{4cm} \rule[-.5cm]{4cm}{0cm}}
  \includegraphics[width=1.3\columnwidth]{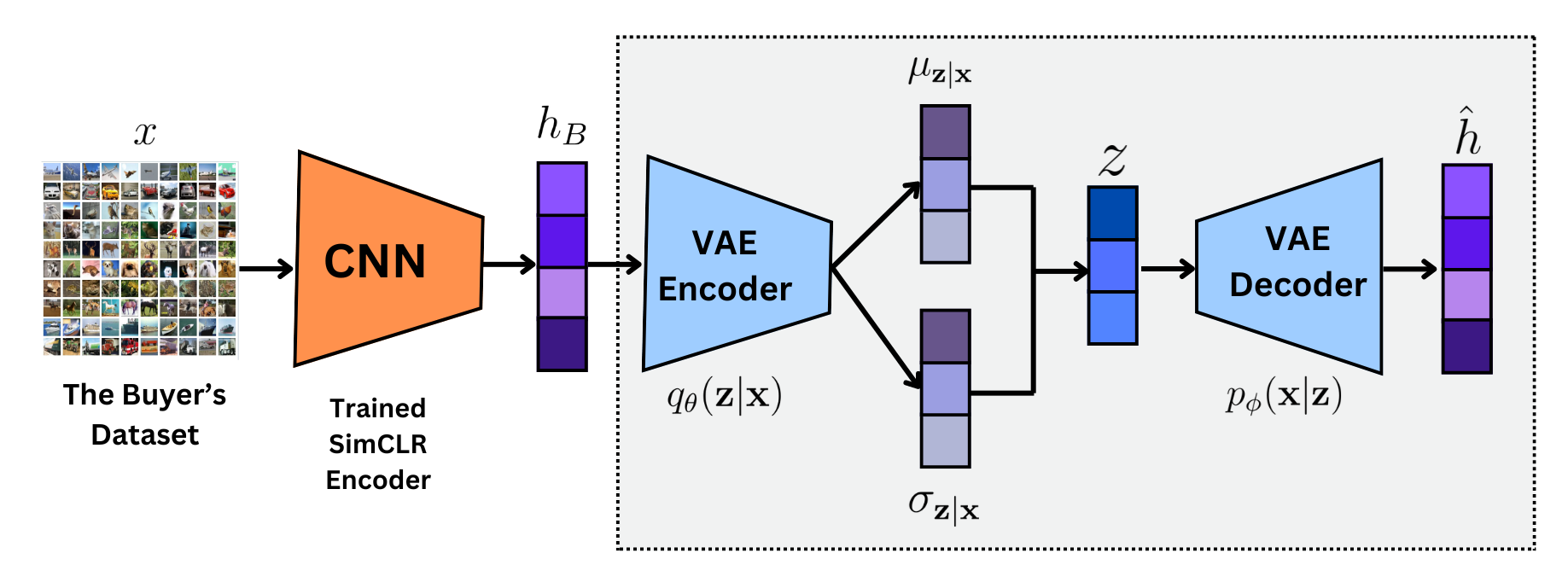}
  \caption{VAE training at the buyer side}
  \label{vae_buyer}
\end{figure*}
\item 
 The buyer shares its trained deep learning model, which represents the mapping function and consists of a concatenation of the SimCLR backbone and the VAE encoder, with the seller entities. Each seller, in turn, as depicted in Fig.~\ref{seller_inf} conducts an inference process utilizing the shared model with a random subset of their datasets to compute the corresponding latent variables as representations of its dataset, denoted as $\mathbf{Z}_{S_i}$ for seller-$i$. 
\item We assume that all entities within the data marketplace are honest, meaning no entity, especially the sellers, intends to deceive the buyer or use malicious behavior to influence a purchase. However, we also assume that each seller wants to keep their dataset, or its representations, private before any transactions. In this step, each seller is asked to send the mean and covariance of the representation of its dataset to the buyer. 
% we need estimators that are not only accurate with respect to the underlying distribution, but also protect the privacy of the individuals represented in the sample. Differential privacy requires adding random noise to some stage of the estimation procedure, and this noise might increase the error of the final estimate.
To make the representations private even after collecting the mean and covariance by the buyer, each seller adds Gaussian noise to the representations as differential privacy with Gaussian mechanism.

Suppose the seller has $n$ representation vectors $\mathbf{z}_j\in \mathbb{R}^d$ (i.e., $\mathbf{Z}_{S_i}\in\mathbb{R}^{n\times d}$) where each vector $\mathbf{z}_j$ is bounded in terms of $\ell_2$-norm, i.e., $\norm{\mathbf{z}_j}_2\le R $. Then the $\ell_2$-sensitivity of mean function, $\mu=\frac{1}{n}\sum_{j=1}^n \mathbf{z}_j$, is bounded by  $\Delta_{\mu}=\frac{2R}{n}$ which 
 measures the maximum change in the mean when a single vector $\mathbf{z}_i$ is replaced by any other vector. Likewise, one can verify that the sensitivity of the covariance function $\Sigma = \frac{1}{n-1}\sum_{j=1}^n (\mathbf{z}_j-\mu)(\mathbf{z}_j-\mu)^T$ is bounded by $\Delta_{\Sigma}=\frac{4R^2}{n}+\frac{8R^2}{n^2}$
 (see Appendix).  For a certain $\epsilon,\delta\in(0,1)$, to achieve $(\epsilon,\delta)$-differential privacy in mean and covariance computation simultaneously, seller-$i$ should add random Gaussian noise $\mathbf{N}_{i}\sim \mathcal{N}(0,\sigma^2 I)$ to its representation $\mathbf{Z}_{S_i}$, where $\sigma={\frac{\Delta_{\Sigma}}{\epsilon }\sqrt{2\ln{(1.25/\delta)}}}$. Therefore, seller-$i$ calculates the mean and covariance of its noisy data representations, denoted by $\mu_{S_i}$ and $\Sigma_{{S_i}}$ respectively.
 
\item The sellers return the calculated statistics to the buyer for further calculations. Note that the subset of each seller's dataset for inference is chosen uniformly at random from their dataset, and its size should be fixed for all sellers. In addition, the choice of $\epsilon$ and $\delta$ for the differential privacy depends on the desired balance between privacy and score of the data involved.

% the statistical information of the corresponding latent variable distribution, i.e., the mean and the covariance. 
\begin{figure*}
  \centering
  % \fbox{\rule[-.5cm]{0cm}{4cm} \rule[-.5cm]{4cm}{0cm}}
  \includegraphics[width=1.3\columnwidth]{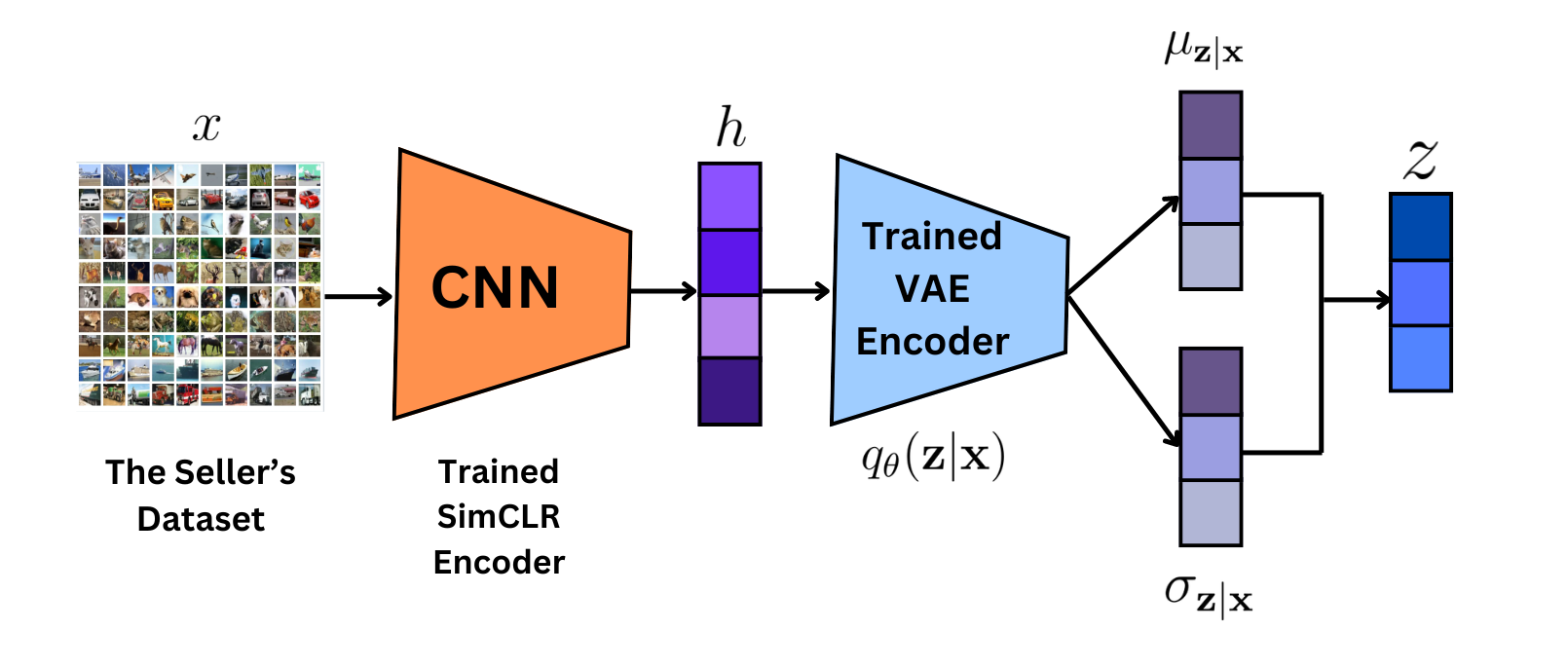}
  \caption{Inference of the trained model at the seller side }
  \label{seller_inf}
\end{figure*}
% , comparable to the size of the buyer's dataset.
% To combat inference attacks against data from each seller before it is sent to the buyer, sellers would add noise to their raw data, compute the noisy mean and covariance, and then send these noisy statistics to the buyer.
\item Since the distribution of the representations of all datasets, even after the addition of noise, is Gaussian, it becomes straightforward for the buyer to apply the Wasserstein distance metric to calculate the gain that each seller's dataset might contribute. More precisely, for seller-$i$ the distance is calculated as
\begin{align}
  \label{wasser_dist}
     W_2(\tilde{P}_B,\tilde{P}_{S_i})=\big(&\norm{\mu_B-\mu_{S_i}}^2 + \tr(\Sigma_B)+\tr(\Sigma_{S_i})\\\nonumber
    &-2\tr\big( (\Sigma_B^{1/2}\Sigma_{S_i}\Sigma_B^{1/2})^{1/2}\big)\big)^{1/2},
\end{align}
where $\mu_B$ and $\Sigma_B$ are the corresponding mean and covariance of the representations of the buyer's dataset achieved by performing inference with the trained model. Additionally, $\tilde{P}_B$ and $\tilde{P}_{S_i}$ denote the distributions of the buyer's and seller-$i$'s representations, respectively.
% Note that the buyer also performs an inference of the model to calculate the corresponding statistics of its current dataset.

% It is worth noting that Maximum Mean Discrepancy (MMD) \cite{gretton2012kernel} is an alternative method for computing the distance between two distributions. 
% However, it only measures the difference in expectations of  a reproducing kernel applied to the data, and it is only sensitive to the mean.

% In addition, the Wasserstein distance can be more scalable and efficient compared to MMD, especially in high-dimensional spaces.
\item  After computing the mutual distances between the buyer’s dataset and each seller’s dataset,  the buyer can make a decision based on the result. If the buyer selects the seller’s data with the greatest distance or dissimilarity, they can better cover the target population; if they select the dataset with the lower dissimilarity, they can enrich their existing dataset in the areas they already have data. Therefore, based on the calculated distance, the buyer can purchase data from the seller that best meets their needs.
% to purchase the data with the greatest dissimilarity, indicating the highest potential value for enhancing the buyer’s dataset. More precisely, 
% \begin{align*}
%     S^*=\arg\max_{i} W_2(\tilde{P}_B,\tilde{P}_{S_i}),
% \end{align*}
% where $S^*$ is the selected seller for the purchase.
\end{itemize}
The $\mathsf{PriArTa}$ method offers three significant advantages: it evaluates entire datasets instead of individual data points, which makes it computationally efficient even at large scales. It is also robust to common data transformations, ensuring that the assigned value of a dataset remains almost consistent even when modified (e.g., through resizing, cropping, or color adjustments). Additionally, $\mathsf{PriArTa}$ enables buyers to assess the value of datasets without direct access to raw data, protecting sellers' privacy by allowing them to share only preprocessed and masked information.

% it requires choosing a kernel, which introduces additional hyperparameters that need tuning. The Wasserstein distance provides a more straightforward and robust measure.

\section{Experiments and Results}
In this section, we introduce the implementation details of $\mathsf{PriArTa}$, followed by the presentation of experimental results to demonstrate the performance of the proposed scheme. All experiments are simulated on a single machine using 
Intel(R) Core(TM) i9-7940X CPU @ 3.10GHz and NVIDIA Quadro P2000 with 5GB GDDR5 memory.
% Intel(R) Xeon(R) CPU @ 2.30GHz and an NVIDIA Tesla T4 GPU with 15 GB of memory, accessed through Google Colab Pro.
\subsection{Implementation Details}
Our experiments are conducted on the CIFAR-10 \cite{krizhevsky2009learning} and STL-10 \cite{coates2011analysis} datasets, which consists of 60,000 (32x32) and 13,000 (96x96) labeled color images across 10 classes, respectively.
To simulate the buyer's and sellers' datasets with varying levels of similarity to the buyer’s data, we first split the dataset into several subsets according to predefined label distributions.  In other words, the buyer and all sellers have subsets of the original datasets, CIFAR10 or STL10, each with a different distribution of classes, leading to imbalanced datasets.

Some of the subsets are then preprocessed using various transformations, such as random flipping, rotation, color jittering, and Gaussian blurring, to generate additional datasets for other sellers. However, these are not new datasets and will not provide additional learning information to the buyer after purchase, as they are merely augmented versions of the existing sellers' or the buyer's datasets. 

To learn meaningful representations from the data, we employ the SimCLR self-supervised learning approach, as the proposed scheme. We utilize the ResNet-18\cite{he2016deep} as the backbone architecture and train a SimCLR model on the buyer's dataset using the PyTorch Lightning framework \cite{falcon2019pytorchlightning}. The model is trained for 200/400 epochs with a batch size of 128/64, using the NTXentLoss as the contrastive loss function. The learned representations from the SimCLR model serve as the input to the subsequent VAE model.
The VAE model consists of an encoder and a decoder network. The encoder takes the SimCLR representations as input and maps them to a latent space of dimension 64/32. The encoder network consists of fully connected layers with LeakyReLU activations and outputs the mean and log-variance of the latent distribution. The decoder network receives the sampled latent vector and reconstructs the original SimCLR representations through a series of fully connected layers.

The parameters of the backbone of the SimCLR is frozen and the VAE model is trained on the buyer’s data using the Adam optimizer. The loss function incorporates both the mean squared error (MSE) between the input and reconstructed representations, as well as the Kullback-Leibler (KL) divergence regularizer. 
After training, the combination of the SimCLR backbone and the VAE encoder is shared with other sellers for inference and computing the likelihood of their datasets. More precisely, each seller randomly selects a subset of their dataset of a certain size, then uses the shared model to compute the corresponding latent variables as representations of their datasets. To make the representation private, each seller employs differential privacy using the Gaussian mechanism with parameters $\epsilon = 0.8$ and $\delta = 10^{-5}$.
 Next, each seller returns the empirical mean and the covariance matrix of the noisy representations to the buyer, who computes the Wasserstein distance between the latent distributions of the buyer’s data and the sellers' data using \eqref{wasser_dist}, i.e., the final valuation scores of each dataset.

\begin{figure*}[ht]
    \centering
     \setlength\tabcolsep{-3pt}
    \begin{tabular}{@{}ccc@{}}
        \includegraphics[width=0.56\columnwidth]{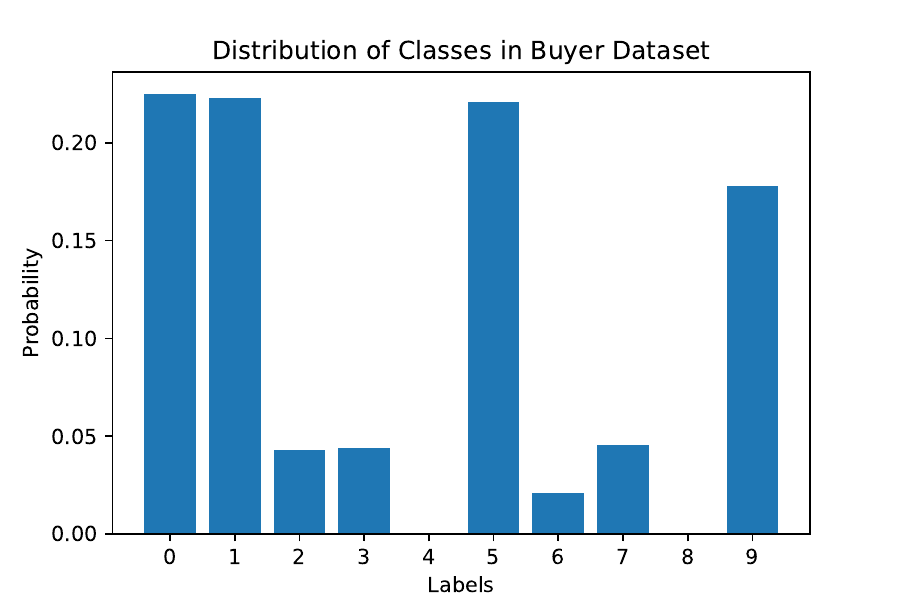}&
        \includegraphics[width=0.56\columnwidth]{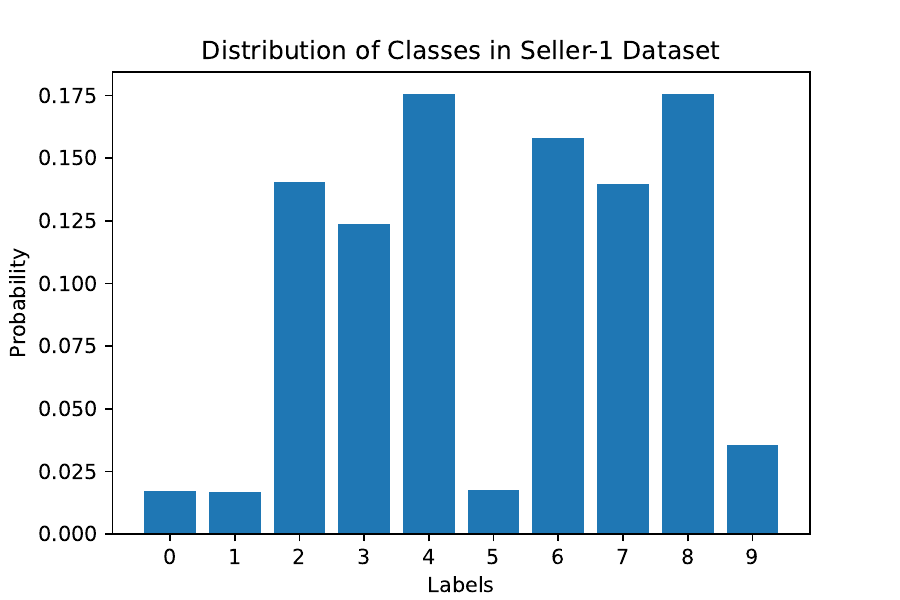} &
        \includegraphics[width=0.56\columnwidth]{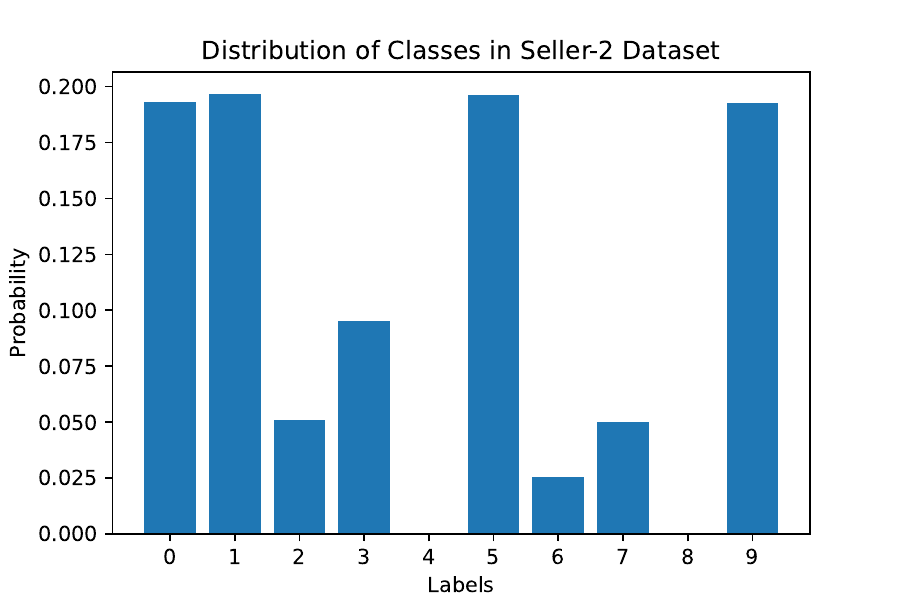} \\
        (a)  & (b) & (c) \\
    \end{tabular}
%     \begin{tabular}{@{}c@{}}
%     \includegraphics[width=0.7\columnwidth]{figs/class_distributions_buy.pdf} \\
%     (a) \\
%     \includegraphics[width=0.7\columnwidth]{figs/class_distributions_s1.pdf} \\
%     (b) \\
%     \includegraphics[width=0.7\columnwidth]{figs/class_distributions_s2.pdf} \\
%     (c) \\
% \end{tabular}
    \caption{The probability distribution of datasets based on their respective classes. (a) represents the distribution of the buyer's dataset, while (b) and (c) represent distributions for seller-1 and seller-2 respectively.}
    \label{fig:dists}
\end{figure*}

\begin{figure}[ht]
    \centering
\begin{tabular}{@{}c@{}}
\vspace{-1mm}
\includegraphics[width=\columnwidth]{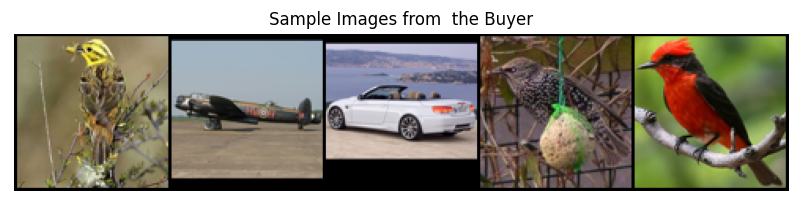} \\
\vspace{-2mm}
\includegraphics[width=\columnwidth]{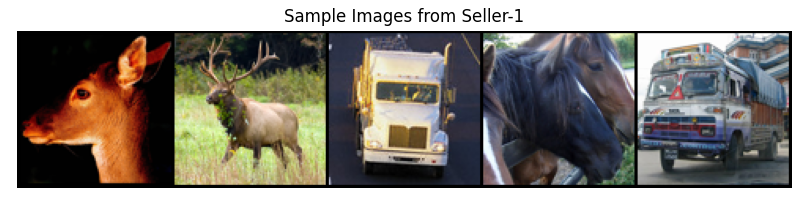} \\
\vspace{-2mm}
\includegraphics[width=\columnwidth]{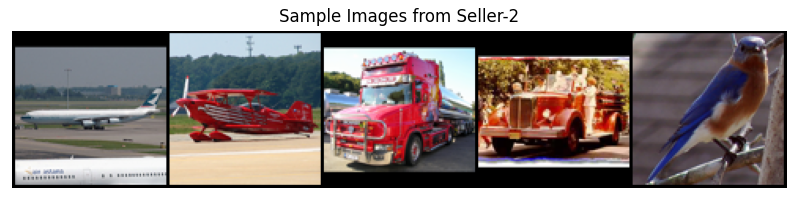} \\
\vspace{-1mm}
\includegraphics[width=\columnwidth]{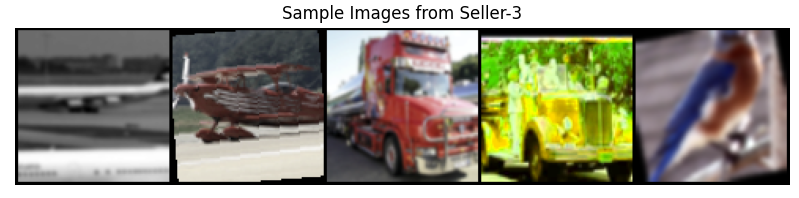} \\
\includegraphics[width=\columnwidth]{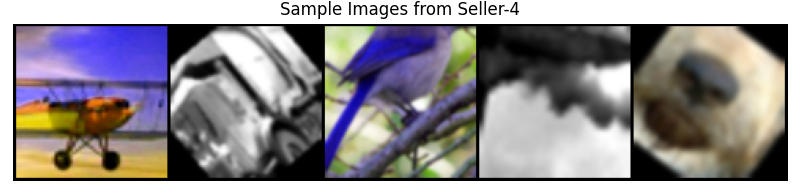} 
\vspace{-1mm}
\end{tabular}
    \caption{Sample images of the buyer's dataset and four sellers'.}
    \label{fig:images}
\end{figure}

\begin{figure}[ht]
  \centering
  % \fbox{\rule[-.5cm]{0cm}{4cm} \rule[-.5cm]{4cm}{0cm}}
  \includegraphics[width=\columnwidth]{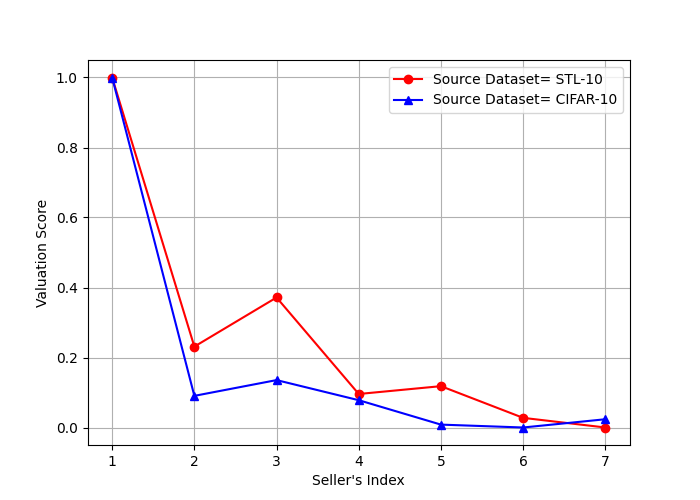}
  \caption{The valuation score of different sellers' datasets, where the underlying datasets are CIFAR-10 and STL-10. }
  \label{score}
\end{figure}

For example, assume that the distributions of different classes in the buyer's dataset and those of two other sellers are illustrated in Fig.~\ref{fig:dists}. In this example, seller-2 has classes that are similar to those in the buyer's dataset, whereas seller-1 covers additional classes, thereby offering greater diversity in the dataset available for purchase by the buyer. 
In addition, we consider five other sellers: seller-3 uses the dataset from seller-2 but applies multiple random transformations to images with varying probabilities. Seller-4 employs the buyer's dataset with multiple random transformations applied at different probabilities. Seller-5 utilizes the buyer's dataset with random flipping exclusively, while Seller-6 applies rotation alone to the same dataset. Lastly, Seller-7 applies only color jittering to the buyer's dataset. Fig.~\ref{fig:images} shows some sample images  of the buyer's dataset and four defined sellers', where the underlying dataset is STL-10. 
Fig.~\ref{score} shows the valuation scores of the aforementioned sellers' datasets achieved by $\mathsf{PriArTa}$, where the underlying dataset are CIFAR-10 and STL-10. 
To emphasize the differences in the Wasserstein distances, we normalize these values. A common approach is to use Min-Max normalization, which scales the values to a fixed range, $[0, 1]$. As shown in Fig.~\ref{score}, if the buyer needs the most diverse dataset, the best option is the dataset from seller-1, as it provides the highest valuation score among all sellers.

\begin{figure}
  \centering
  % \fbox{\rule[-.5cm]{0cm}{4cm} \rule[-.5cm]{4cm}{0cm}}
  \includegraphics[width=\columnwidth]{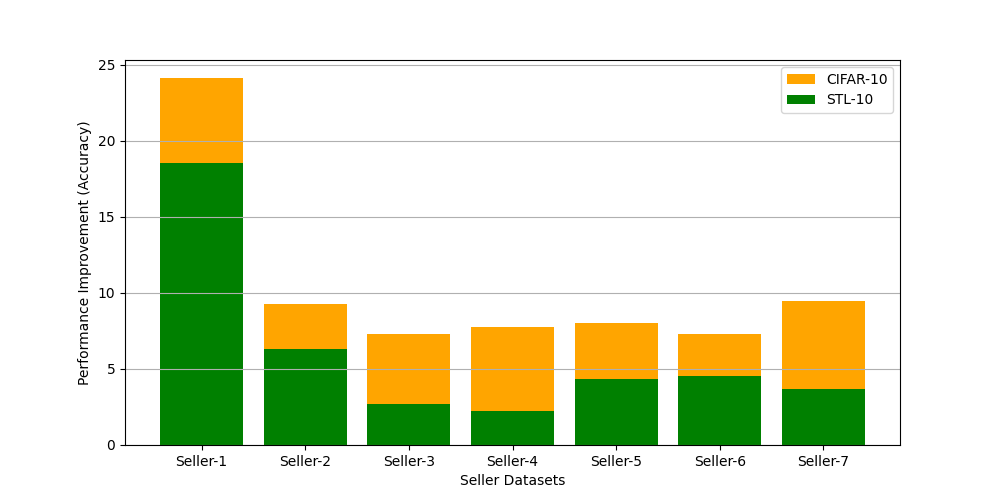}
  \caption{ Performance improvement of the buyer's model with  seller datasets. }
  \label{accuracy}
\end{figure}

The overall goal of the buyer in purchasing data is to perform a task, which in this example, we assume to be classification. Initially, the buyer will train its model on the local dataset. After purchasing a new dataset, it will fine-tune the model. We assume that the VGG-16 network \cite{simonyan2014very} is used for this classification task. Fig.~\ref{accuracy} shows the model's performance improvement in terms of test accuracy based on training on the buyer's dataset and fine-tuning on other sellers' datasets after 30  epochs. As shown, if the buyer purchases the dataset from seller-1, the accuracy improves the most, which is consistent with the results from the proposed method and the valuation scores.
Note that the accuracy improvement from using sellers 4, 5, 6, and 7 is not due to the novelty or provision of new information. Instead, it is a result of the augmentation technique that the buyer can apply independently, without purchasing additional datasets from these sellers.
% TO DO
% imbalanced dataset
% we want to balance the classes in the subset
% The rationale behind this is that a more dissimilar dataset is likely to provide novel and valuable information, which can enhance the performance and robustness of the trained model.
%  task-agnostic data valuation, which aims to assess the value of data without relying on specific downstream tasks.
%  The theoretical advantages of a task-agnostic approach, such as its generalizability across different domains and its ability to capture intrinsic data properties.
%  By focusing on the intrinsic properties of the data rather than task-specific characteristics, task-agnostic methods can be applied to diverse domains, including images, text, audio, and multimodal data. This broad applicability enhances the score of the method in various contexts.
\vspace{-1em}
\section{Conclusion}
In this paper, we present $\mathsf{PriArTa}$, a novel approach for privacy-preserving and augmentation-robust data valuation in data marketplaces. $\mathsf{PriArTa}$ is a task-agnostic method that enables buyers to evaluate the whole dataset of each seller without full access, ensuring privacy and minimizing redundancy in the purchase. Experimental results on real-world image datasets demonstrate the effectiveness of $\mathsf{PriArTa}$ in providing reliable data valuation, even in the presence of sellers who have augmented versions of each other's datasets.
% \section{Acknowledgement}

\nocite{langley00}
\bibliography{references}
\bibliographystyle{mlsys2025}

%%%%%%%%%%%%%%%%%%%%%%%%%%%%%%%%%%%%%%%%%%%%%%%%%%%%%%%%%%%%%%%%%%%%%%%%%%%%%%%
%%%%%%%%%%%%%%%%%%%%%%%%%%%%%%%%%%%%%%%%%%%%%%%%%%%%%%%%%%%%%%%%%%%%%%%%%%%%%%%
% SUPPLEMENTAL CONTENT AS APPENDIX AFTER REFERENCES
%%%%%%%%%%%%%%%%%%%%%%%%%%%%%%%%%%%%%%%%%%%%%%%%%%%%%%%%%%%%%%%%%%%%%%%%%%%%%%%
%%%%%%%%%%%%%%%%%%%%%%%%%%%%%%%%%%%%%%%%%%%%%%%%%%%%%%%%%%%%%%%%%%%%%%%%%%%%%%%
\appendix
\section{Sensitivity of Mean and Covariance Functions}
Suppose we have $n$ vectors $\mathbf{x}_i\in\mathbb{R}^d$, where each vector has a bounded $\ell_2$-norm such that $\norm{\mathbf{x}_i}_2\le R$. We define function $\mu$ that computes the mean of these vectors as $\mu=\frac{1}{n}\sum_{i=1}^n \mathbf{x}_i$. To determine the $\ell_2$-sensitivity of the mean function, we need to consider the effect of changing one vector $\mathbf{x}'_j$ among all vectors. When a single vector is changed, the maximum change in the mean is bounded as
\begin{align}
    \norm{\mu'-\mu}_2= \norm{\frac{1}{n}(\mathbf{x}'_j-\mathbf{x}_j)}_2\le\frac{2R}{n},
\end{align}
where $\mu'$ is the mean vector after the change, and the last inequality holds because of the assumption that all vectors are bounded within a sphere of radius $R$.

To calculate the change in the covariance $\Sigma=\frac{1}{n-1}\sum_{i=1}^n (\mathbf{x}_i-\mu)(\mathbf{x}_i-\mu)^T$  due to the change in a single vector, we should consider the worst case change which occurs when we replace a vector that was $R$ away from the mean in one direction with a vector that is $R$ away from the mean in the opposite direction. Let us denote the original vector by $\mathbf{x}_j=\mu+R\mathbf{u}$ and the replaced one by $\mathbf{x}'_j=\mu-R\mathbf{u}$, where $\mathbf{u}$ is a unit vector in some direction.  The changes in the covariance matrix when we replace $\mathbf{x}_j$ with $\mathbf{x}'_j$ is 
\begin{align*}
    \Delta\Sigma = \frac{1}{n-1}\bigg((\mathbf{x}'_j-\mu')(\mathbf{x}'_j-\mu')^T-(\mathbf{x}_j-\mu)(\mathbf{x}_j-\mu)^T \\
    +\sum_{i=1,i\ne j}^n \big((\mathbf{x}_i-\mu')(\mathbf{x}_i-\mu')^T-(\mathbf{x}_i-\mu)(\mathbf{x}_i-\mu)^T\big) \bigg),
\end{align*}
where $\mu'$ is the new mean after replacing $\mathbf{x}_j$ with $\mathbf{x}'_j$ and we have 
$\mu'= \mu + (\mathbf{x}'_j-\mathbf{x}_j)/n= \mu-2R\mathbf{u}/n$. In addition, we have
\begin{align*}
    \mathbf{x}'_j-\mu'=(\mathbf{x}'_j-\mu)+(\mu-\mu')=-R\mathbf{u}+2R\mathbf{u}/n.
\end{align*}
Substituting these back into the $\Delta\Sigma$
\begin{align*}
    \Delta\Sigma =&\frac{1}{n-1}\bigg( R^2\mathbf{u}\mathbf{u}^T(1-\frac{2}{n})^2-R^2\mathbf{u}\mathbf{u}^T\\
    &+\sum_{i=1,i\ne j}^n\big( \frac{2R}{n}((\mathbf{x}_i-\mu)\mathbf{u}^T+\mathbf{u}(\mathbf{x}_i-\mu)^T)+\frac{4R^2}{n^2}\mathbf{u}\mathbf{u}^T\big)\bigg).
\end{align*}
Using triangle inequality properties of the Frobenius norm we have
\begin{align*}
    \norm{\Delta\Sigma}_F\le& \frac{1}{n-1}\bigg( \frac{4R^2}{n}|(\frac{1}{n}-1)|\norm{\mathbf{u}\mathbf{u}^T}_F \\
    &+ \sum_{i=1,i\ne j}^n \big(\frac{4R}{n}\norm{(\mathbf{x}_i-\mu)\mathbf{u}^T}_F+\frac{4R^2}{n^2}\norm{\mathbf{u}\mathbf{u}^T}_F\big)\bigg)\\
    &\le \frac{4R^2}{n}+\frac{8R^2}{n^2}.
\end{align*}
% The maximum absolute value of any entry in $\mathbf{u}\mathbf{u}^T$ is 1 occurs when $\mathbf{u}$ is the standard basis vector. 
% Note that this is not a tight bound, and further analysis may lead to a tighter bound.

% %%%%%%%%%%%%%%%%%%%%%%%%%%%%%%%%%%%%%%%%%%%%%%%%%%%%%%%%%%%%%%%%%%%%%%%%%%%%%%%
%%%%%%%%%%%%%%%%%%%%%%%%%%%%%%%%%%%%%%%%%%%%%%%%%%%%%%%%%%%%%%%%%%%%%%%%%%%%%%%

\end{document}